\documentclass[letterpaper]{article}
\usepackage{aaai}
\usepackage{times}
\usepackage{helvet}
\usepackage{courier}
\usepackage{amsmath}
\usepackage{booktabs}
\usepackage{xcolor}
\usepackage{threeparttable}
\usepackage{graphicx}
\usepackage{subfig}
\usepackage{makecell}
\begin{document}
%
\title{Cross-domain feature disentanglement for interpretable modeling of tumor microenvironment impact on drug response}
\author{Jia Zhai$^1$, Hui Liu$^{1*}$\\
$^1$College of Computer and Information Engineering, Nanjing
Tech University, Nanjing, 211816, China.\\
$^{*}$Corresponding author. E-mail(s): hliu@njtech.edu.cn\\
}
\maketitle
\begin{abstract}
\begin{quote}
High-throughput screening technology has facilitated the generation of large-scale drug responses across hundreds of cancer cell lines. However, there exists significant discrepancy between in vitro cell lines and actual tumors in vivo in terms of their response to drug treatments, because of tumors comprise of complex cellular compositions and histopathology structure, known as tumor microenvironment (TME), which greatly influences the drug cytotoxicity against tumor cells. To date, no study has focused on modeling the impact of the TME on clinical drug response. This paper proposed a domain adaptation network for feature disentanglement to separate representations of cancer cells and TME of a tumor in patients. Two denoising autoencoders were separately used to extract features from cell lines (source domain) and tumors (target domain) for partial domain alignment and feature decoupling. The specific encoder was enforced to extract information only about TME. Moreover, to ensure generalizability to novel drugs, we applied a graph attention network to learn the latent representation of drugs, allowing us to linearly model the drug perturbation on cellular state in latent space. We calibrated our model on a benchmark dataset and demonstrated its superior performance in predicting clinical drug response and dissecting the influence of the TME on drug efficacy. 
\end{quote}
\end{abstract}

\section{Introduction}
The advancement of high-throughput drug screening technologies has facilitated the generation of large-scale drug response datasets, such as the GDSC \cite{2013Genomics,2016A} and CCLE \cite{2012The} databases. These repositories contain a wealth of information on bulk RNA-seq and drug sensitivity spanning thousands of tumor cell lines and hundreds of drug molecules. While a couple of studies have proposed to predict the drug responses of cell lines in vitro \cite{2014Clinical}, with considerable predictive power, the ultimate goal in clinical practice is to identify the response of tumors in vivo to drug treatment and provide actionable options for personalized medicine. However, existing computational methods with good performance for in vitro drug sensitivity often perform poorly in predicting in vivo drug response \cite{2020Assessment}. The discrepancy can be largely attributed to the significant difference between cell line and actual tumors in patients. In contrast to cultured cells in vitro, tumor cells in patients always interact with their surrounding environment to form a specific tumor structure having complex cell composition and carcinogenic functions, known as the tumor microenvironment (TME) \cite{HANAHAN2012309}. The TME plays a pivotal role in tumor proliferation, progression, and metastasis, and also exerts a significant impact on drug efficacy \cite{2017Tumor}. Consequently, the straightforward application of in vitro prediction models to in vivo situations often results in inferior performance. Therefore, there is an urgent need for the development of computational methods that can accurately predict actual tumor drug response.
\vspace{-0.5cm}
\begin{figure}[htbp]
    \centering
    \centerline{\includegraphics[width=8cm]{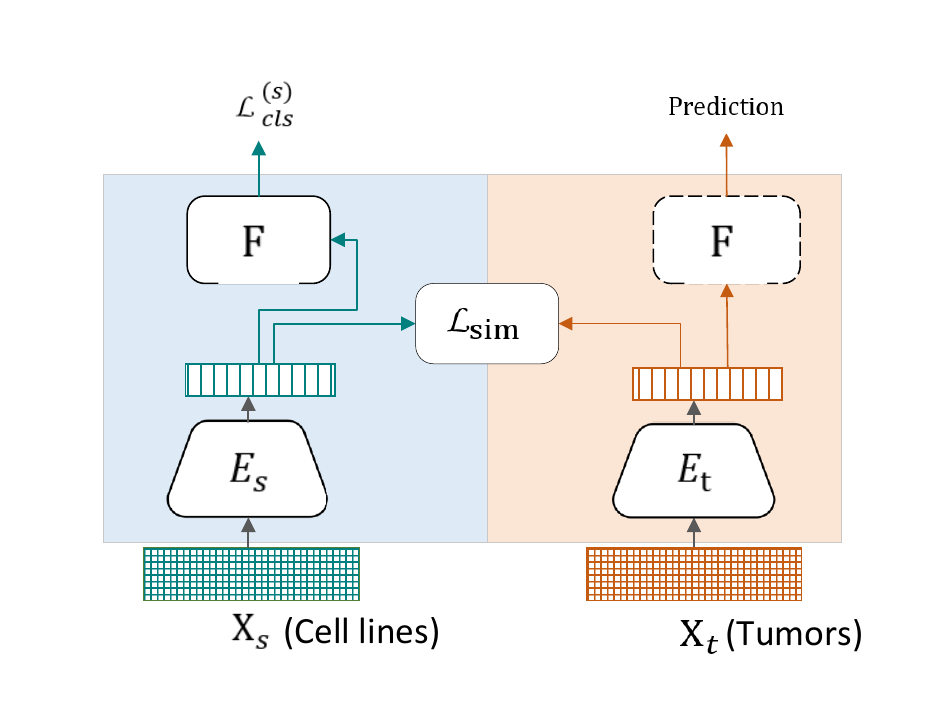}}
    \caption{Illustrative flowchart of the conventional domain adaptation for clinical drug response prediction
} \label{fig:baseline}
\end{figure}

Some studies have been proposed to predict the response of cancer patients to specific drug treatment. For instance, Geeleher et al \cite{2014Clinical} firstly removed batch effects between cell lines and tumors, and then utilized the cell line biomarkers to predict tumor drug response. Some works transformed transcriptomic profiles into signaling pathway activities for interpretable modeling of drug response \cite{chawla2022gene}. However, these methods have not considered the discrepancy in data distribution between in vitro and in vivo expression profiles. To mitigate this issue, transfer learning has been employed to predict in vivo drug response. For example, Mourragui et al. \cite{2019PRECISE} leveraged domain adaptation to extract common features between preclinical and patient tumors for drug response prediction. The TRANSACT model \cite{mourragui2021predicting} employed kernel methods to capture common linear and nonlinear molecular processes in vitro and in vivo that is helpful for drug response prediction. AITL \cite{sharifi2020aitl} utilized adversarial domain adaptation and multi-task learning to solve the domain bias problem. He et al. \cite{he2022context} applied domain separation network to capture common and domain-specific features from cell line and tumor transcriptomic profiles, which separated the drug response signals from confounding factors and achieved significant progress in predicting tumor drug response. Sharifi et al. \cite{sharifi2021out} adopted multi-source domain adaptation to align features extracted from multiple cell lines and patients for drug response. TCRP \cite{ma2021few} exploited few-shot learning to fine-tune a pre-trained model on cell line transcriptomic data using only a few patient drug response samples. The domain adaptation methods focus on extracting domain-invariant features between the source domain (cell lines or xenograft data) and the target domain (actual tumor), thereby transferring the knowledge learned in source domain with sufficient training data to the target domain, as shown in Figure \ref{fig:baseline}. However, the actual tumor drug response in vivo is influenced by many factors, including the tumor microenvironment, which is a crucial factor that is often overlooked by existing models. As a result, simply aligning the transcriptomic features between tumors and in vitro cell lines through domain adaptation could just yield limited performance.

In this paper, we proposed Drug2TME, a deep learning framework devoted to estimate the impact of the TME on clinical drug response. We employed a domain separation network to decouple the tumor feature into two parts: cancer cells and TME. In particular, two denoising autoencoders, referred to as the common encoder and specific encoder, were used to extract two orthogonal features from tumor transcriptomic data (target domain). By aligning the cancer cell feature to cell line (source domain) via domain adaptation, the specific encoder was enforced to extract information only about TME. Moreover, for model generalizability to novel drugs, we applied a graph attention network to learn the latent representation of drugs, allowing us to linearly model the drug perturbation on cellular state in latent space. We evaluated our model on both patient-derived tumor xenograft cell (PDTC) and tumor patient drug responses, and the experimental results demonstrate its superior performance in predicting clinical drug response and dissecting the influence of the TME on drug efficacy. 

We believe this work has at least three contributions as below:
\begin{itemize}
    \item As far as our knowledge, this is the first deep learning framework proposed to model the impact of tumor microenvironment on drug response.
    \item Rather than mechanistic modeling, we leverage feature disentanglement and partial domain adaptation to dissect the information of tumor microenvironment from transcriptomic profiles, which enable us to independently quantify its role in the response to drug treatment.
    \item We evaluated Drug2TME on both patient-derived tumor xenografts (PDTX) and patients samples, and the experimental results demonstrated that the proposed model can effectively improve the prediction performance of drug response.
\end{itemize}

\section{Related Works}
\subsection{Domain Adaptation}
Domain adaptation is a subfield of transfer learning that focuses on improving the model generalizability when the source and target domains have different data distributions. By uncovering the domain-invariant feature, domain adaptation enables the transfer of knowledge learned from a source domain with abundant labeled data to a target domain that may lack sufficient labeled data. One-step deep domain adaptation methods \cite{wang2018deep} have been widely applied in computer vision and have achieved remarkable success, which are mostly based on feature alignment, including discrepancy-based methods \cite{tzeng2019deep,long2015learning,long2017deep,pan2019transferrable,lee2019sliced}, adversarial-based methods \cite{ganin2015unsupervised,zhao2018adversarial,long2018conditional,chen2019blending,cao2019learning,you2019universal}, and reconstruction-based methods \cite{ghifary2016deep,bousmalis2016domain}. In this work, we applied domain adaptation to align the features between cell lines and tumors.

\subsection{Feature Disentanglement}
Feature disentanglement is a unsupervised learning technique that separates the underlying factors for more robust and interpretable models \cite{bengio2013representation,ridgeway2018learning}. 
Feature disentanglement based methods for solving DA problem have been widely used in computer vision \cite{bousmalis2016domain,cai2019learning,peng2019domain,saito2019strong}. Inspired by the feature disentanglement, we decouple the tumor transcriptomic profile into two independent features, tumor cells and tumor microenvironment, which thereby allow us to estimate their influences on drug response separately.

\subsection{Drug Response Prediction}
As accurate prediction of clinical drug response is indeed a crucial step towards precision medicine, several methods have been proposed to predict drug response, including multi-task learning \cite{sharifi2020aitl}, adversarial learning \cite{dincer2020adversarial}, and few-shot learning \cite{ma2021few}. Our method is mostly related to CODE-AE \cite{he2022context} that combined domain adaptation with feature decoupling to predict the response of cancer patients to specific drug.

\section{Methods}

\subsection{Overview of Drug2TME framework}
\begin{figure*}[!htbp]
    \centering
    \centerline{\includegraphics[width=15cm]{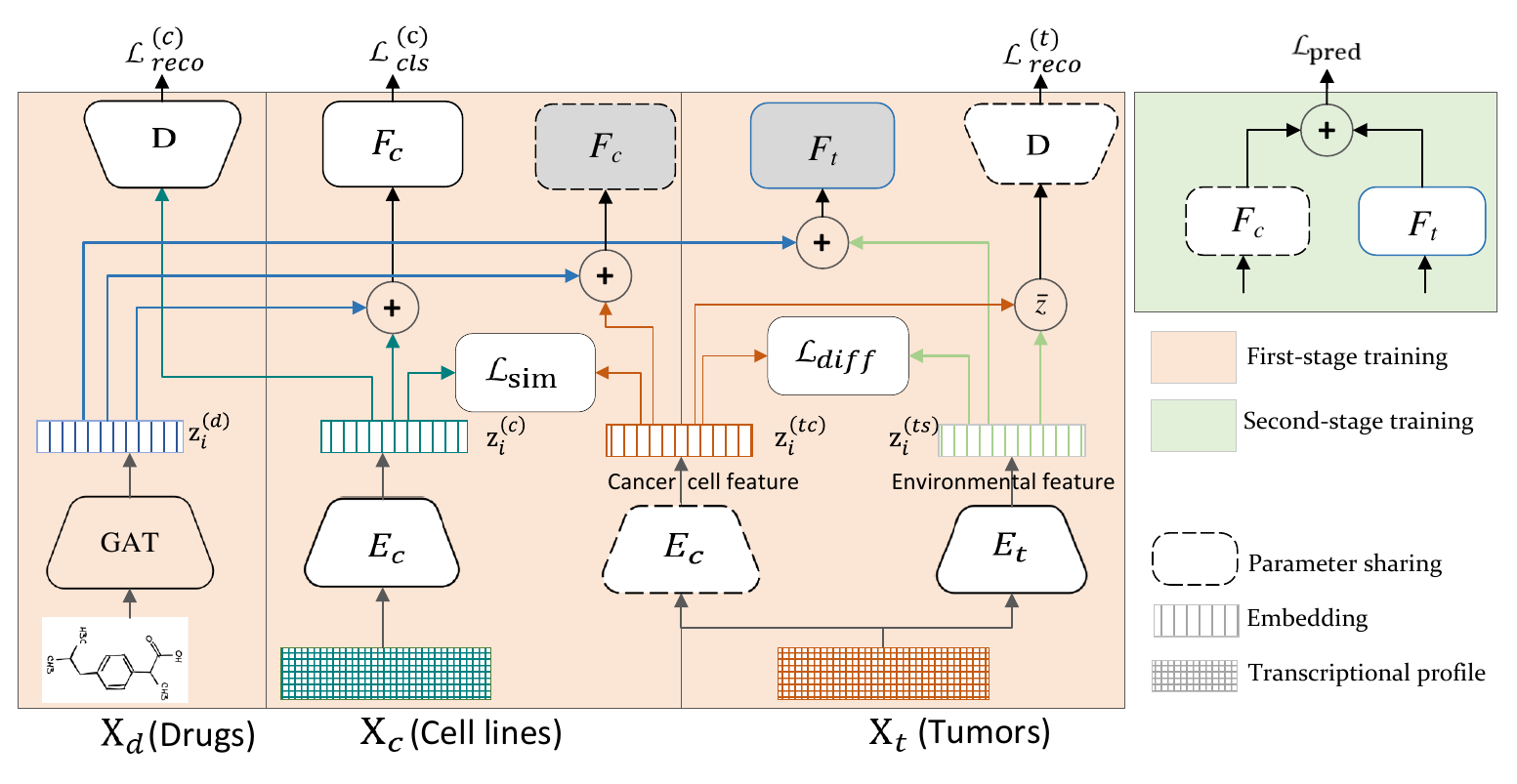}}
    \caption{Illustrative architecture of the proposed Drug2TME method for modeling the impact of tumor microenvironment on drug response
} \label{fig:framework}
\end{figure*}

As a tumor was composed of two parts: cancer cells and the tumor microenvironment, the tumor bulk transcriptomic data was actually a mixture of signals from these two components. To explicitly model the impact of tumor microenvironment on drug efficacy, we proposed a cross-domain feature disentanglement framework, as shown in Figure \ref{fig:framework}. A domain separation network was introduced to separately encode the transcriptomic features that respectively correspond to cancer cells and TME, referred to as CanCell feature and TME feature. Next, the CanCell feature was aligned to the cell line feature via domain adaptation and compulsively orthogonal to TME feature, so that the TME feature was separated from cancer cells and contains only the information about tumor microenvironment. In the first training stage, the domain separation network and domain adaption were simultaneously trained so that a source domain classifier $F_c$ is fit to predict cell line drug response. In the second stage, all trainable parameters were frozen except another predictor regarding TME feature $F_t$ was trained so that the summation of $F_c$ and $F_t$ on target domain was fit to the cancer drug response. Moreover, to improve model generalizability to unseen drugs, we introduced a pre-trained drug encoder and fused it with gene expression features to predict the response of cells and tumor patients to new drugs.
\subsection{Cell line and drug encoders}
To encode the cell line feature, a denoising autoencoder was used to map their transcriptional data to low dimensional representations in a latent space. The expression profiles standing for cellular state served as the input of the encoder, and the decoder endeavored to recover them. Formally, assume that we had the expression profiles of cell lines $X_c=\{x_i^{(c)}\}_{i=1}^{N_c}$, in which $N_c$ represented the number of cell lines. The encoder $E_c$ mapped the cell line expression profiles $x_i^{(c)}$ to a latent representation $z_i^{(c)}$, while the decoder $D$ tried to recover the input data, $\hat{x}_i^{(c)}={D}({E_c}(x_i^{(c)})$. The mean squared error was used as the reconstruction loss of the autoencoder:
\begin{equation}
    \mathcal{L}^{(c)}_{reco}=\frac{1}{N_c}\sum_{i=1}^{N_c}\left\|x^{(c)}_i-\hat{x}^{(c)}_i\right\|^2
\end{equation}
The encoder and decoder were fully or densely connected neural networks with rectified linear unit (ReLU) activation between layers. The dimension of $z$ was in effect the network information bottleneck.

To model the phenotypic perturbation of drugs acting on cells, we used a encoder to map drug molecules to the same latent space. Specifically, the SMILE of the drug was converted into a graph, which was fed into a graph attention network (GAT) to obtain the drug feature. Denote by $X_d=\{x_i^{(d)}\}_{i=1}^{N_d}$ , the molecular graphs converted from drug SMILES descriptors, in which ${N_d}$ was the number of drugs. The GAT encoder mapped the drug $x_i^{(d)}$ to the representation $z_i^{(d)}$ in the latent space same to mapped transcriptional data. In our practice, we used the GAT encoder pre-trained with attention-wise masked graph contrastive learning in recent study \cite{liu2022attention}.  Once we mapped the expression profiles and drug perturbations to the same latent space, we could formulate the drug efficacy by following the linear additivity model like chemCPA \cite{hetzel2022predicting}. Compared with one model for one drug methods, our model could predict the response of any combination of cell line drugs and patient drug combinations.

\subsection{Domain separation for tumor feature disentanglement}
A domain separation network was used to decouple the tumor expression profiles into two independent features. Denote by  $X_t=\{x_i^{(t)}\}_{i=1}^{N_t}$ the expression profiles of tumors, where ${N_t}$ was the number of tumor samples. Two encoders, the common encoder $E_{c}$ (shared parameters with cell line encoder) and private encoder $E_{t}$, took the expression profiles as input and outputted two embeddings, $z_i^{(tc)}$ and $z_i^{(ts)}$ respectively. For feature disentanglement, we required $z_i^{(tc)}$ and $z_i^{(ts)}$ were orthogonal to each other so that each embedding did not contain any information about each other. For this purpose, we defined the loss function as below:
\begin{equation}
\mathcal{L}_{diff}=\sum_{i=1}^{N_t}\left\|z_i^{(tc)}.z_i^{(ts)}\right\|_F^2,
\end{equation}
where $\left\|.\right\|_F^2$ was the squared Frobenius norm.

Meanwhile, we required that the combination of $z_i^{(tc)}$ and $z_i^{(ts)}$ contains enough information to recover the original signal. In practice, both the target domain and the source domain transcriptional profiles were reconstructed by a shared decoder, so we just computed the mean value $\bar{z}_i^{(t)}= (z_i^{(tc)}+z_i^{(ts)})/2$ instead of concatenation and then took it as the input of the decoder $D$. We required that $\hat{x}_i^{(t)}={D}(\bar{z}_i^{(t)})$ was as close to ${x}_i^{(t)}$ as possible. Therefore, we defined the reconstruction loss from decoupled  tumor feature as below:
\begin{equation}
    \mathcal{L}_{reco}^{(t)}=\frac{1}{N_t}\sum_{i=1}^{N_t}\left\|x_i^{(t)}-\hat{x}_i^{(t)}\right\|^2
\end{equation}

\subsection{Partial domain adaptation}
To formulate the tumor microenvironment feature, we performed partial domain adaptation between cell line and tumor expression profiles. Specially, the embedding $z_i^{(tc)}$ extracted from tumor expression profile was aligned to the cell line embedding $z_i^{(c)}$, such that the common encoder $E_{c}$ was enforced to extract the common feature between tumor and cell line. It is of importance to note that because the domain separation network made the decoupled tumor features orthogonal, the embedding $z_i^{(ts)}$ encoded by the private encoder $E_{t}$ captured all the information except for tumor cells, which was referred to as the TME feature.

We tested two typical methods for domain adaptation, maximum mean discrepancy (MMD) and adversarial learning. The MMD method measured the distance between the means of the two distributions in a reproducing kernel Hilbert space (RKHS), and was widely used to align the distributions of the source and target domains. Given the Gaussian kernel function $k()$, we defined the MMD loss function as below:
\begin{align}
          \mathcal{L}_{sim} &= \frac{1}{N_c^2}\sum_{i,j=1}^{N_c}k\left(z_i^{(c)},z_i^{(c)}\right)+\frac{1} {N_t^2}\sum_{j=1}^{N_t}k\left(z_j^{(tc)},z_j^{(tc)}\right) \nonumber \\
          & -\frac{2}{N_cN_t}\sum_{i,j=1}^{N_c,N_t}k\left(z_i^{(c)},z_j^{(tc)}\right)
\end{align}

For adversarial domain adaptation, we adopted the domain-adversarial neural network (DANN) \cite{ganin2015unsupervised}. For this purpose, a domain classifier and a gradient reversal layer were applied to the extracted features from both cell line and tumor expression profiles. The domain classifier tried to maximize its accuracy in distinguishing the feature domains, while the feature encoder was trained to confuse the domain classifier as much as possible. The adversarial training was implemented via the gradient reversal layer. Denote by $d_i$ and $\hat{d}_i$ the true and predicted domain label (cell line or tumor) of $z_i^{(c)}$ and $z_i^{(tc)}$, the loss function of the domain classifier was defined as
\begin{equation}
    \mathcal{L}_{sim}=\sum_{i=1}^{N_{c}+N_{t}} (d_{i}\log\hat{d}_{i}+(1-d_{i})\log{(1-\hat{d}_{i})})
\end{equation}

\subsection{Interpretable modeling of tumor drug response}
When the expression profiles and drugs were mapped to the same latent space, the drug perturbation on cell viability was often assumed to follow linear additivity in the latent space so that interpretable model could be established \cite{hetzel2022predicting}. Given the cell line embedding $z_i^{(c)}$ and the drug embedding $z_j^{(d)}$, their linear summation was fed into a fully connected network to predict the effect of drugs on cell viability. Denote by $F_c(\cdot)$ the predictor, we got the predicted drug response $\hat{y}_{ij}=F_c(z_i^{(c)}+z_j^{(d)})$ upon drug $j$ exerted to cell line $i$. Let $y_{ij}$ was the actual response, we used cross-entropy loss to optimize the predictor:
\begin{equation}
\mathcal{L}_{cls}^{(c)}=-\sum_{i=1}^{N_{c}}\sum_{j=1}^{N_{d}}[y_{ij}log\hat{y}_{ij}+(1-y_{ij})log(1-\hat{y}_{ij})]
\end{equation}
Upon the drug response prediction task, the common encoder was enforced to extract the features relevant to drug response.

For tumor, we modeled the impact of two partial features on the drug efficacy independently. On the one hand, due to the domain adaptation, the aligned tumor feature $z_i^{(tc)}$ contained domain-invariant information about tumor cells response to drug treatment. Therefore, the trained predictor $F_c$ could be directly used to predict the drug response of tumor cells. Similarly, the summation of tumor embedding $z_i^{(tc)}$ and the drug embedding $z_j^{(d)}$ were fed into $F_c$, and the output $F_c(z_i^{(tc)}+z_j^{(d)})$ represented the partial response to drug treatment. On the other hand, to evaluate the impact of tumor microenvironment on drug efficacy, we introduced another fully connected network $F_t()$, referred to as TME predictor, which took the summation of the TME embedding $z_i^{(ts)}$ and drug embedding $z_j^{(d)}$ as input. As a result, the value $F_t(z_i^{(ts)}+z_j^{(d)})$ represented another partial response to drug treatment. As a whole, we supposed that the drug response of a tumor patient was the linear summation of these two partial response. The linear additivity made our model interpretable to further quantitatively evaluate the role of tumor microenvironment in response to drug. Assume that the true response of patient $i$ to drug $j$ was $r_{ij}$, and the predicted response was $\hat{r}_{ij}$=$F_c(z_i^{(tc)}+z_j^{(d)})+F_t(z_i^{(ts)}+z_j^{(d)})$, we used cross-entropy as the loss function:
\begin{equation}
    \mathcal{L}_{pred}=-\sum_{i=1}^{N_{t}}\sum_{j=1}^{N_{d}}[r_{ij}log\hat{r}_{ij}+(1-r_{ij})log(1-\hat{r}_{ij})]
\end{equation}

The training of our model was consisted of two stages. In the first stage, the expression profiles of both cell lines and tumors, and the drug sensitivity of cell lines, were taken as input to train the autoencoder of the source domain, the feature separation network of the target domain, as well as the domain adaptation and the predictor in the source domain. As such, the full objective of the first stage was defined as below:
\begin{equation}
\mathcal{L}_1=\mathcal{L}^{(c)}_{reco}+\mathcal{L}^{(t)}_{reco}+\mathcal{L}_{sim}+\mathcal{L}_{diff}+\mathcal{L}_{cls}^{(c)}
\end{equation}

In the second stage, the parameters of all encoders and cell line predictors were frozen except for the TME predictor. The TME predictor was trained using the clinical drug responses of tumor patients, namely, the loss function of the second stage $\mathcal{L}_2=\mathcal{L}_{pred}$ was minimized. 

\section{Evaluation Experiments}
\subsection{Construction of Benchmark Datasets}
The drug sensitivity of cell lines was obtained from the Genomics of Drug Sensitivity in Cancer (GDSC) project \cite{2013Genomics,2016A}. The GDSC project contained a wealth of data about the growth responses of various cancer cell lines to a range of therapeutic agents. These responses were quantified using the Area Under the drug response Curve (AUC), which represented the fraction of the total area under the drug response curve between the highest and lowest screening concentrations. For each drug of interest, we first identified all cell lines for which corresponding drug sensitivity data was available, as measured by the AUC metric. Subsequently, we divided the drug sensitivity data for these cell lines into two distinct categories: responsive (sensitive) and non-responsive (resistant). The threshold was determined by calculating the average AUC value for all available cell-line drug sensitivity upon this drug.

For target domain, two different datasets were used: patient-derived tmor xenograft cells (PDTC) from BCaPE biobank \cite{2016Bruna} and tumor patients from TCGA database \cite{weinstein2013cancer}. PDTXs were pre-clinical models of cancer where tumor tissue from a patient was implanted into an immunodeficient mouse, allowing for the growth and study of human tumors in vivo.  The PDTCs in short-term culture retained tumor heterogeneity and tumor microenvironment, making them a useful agents for cancer drug screening and the potential discovery of molecular mechanisms of therapy response. PDTCs were exposed to a range of drugs to obtain model drug response data, including half maximal inhibitory concentration (IC50) and area under the dose response curve of a drug. A total of 1,503 responses from 20 PDTC samples on 94 drugs were collected. Similarly, we binarize the PDTC drug responses based on the AUC values using the threshold provided by BCaPE database. The tumor patient samples came from the TCGA database, with transcriptomic profiles of 9,728 patients collected. The clinical drug responses of the patient samples were collected from clinical metadata, with complete response and partial response patients categorized as responders and patients with clinically progressive and stable diseases marked as non-responders.

The expression profile of each sample was used to compute the pathway activity using the Gene Set Variation Analysis tool (GSVA)  \cite{hanzelmann2013gsva}. The C2 curated gene sets obtained from the Molecular Signatures Database were used for pathway activity calculation. The SMILE descriptors of all drugs were downloaded from the PubChem database \cite{swain2014pubchempy} and then converted into molecular graphs, which were used as input of the GAT encoder.

For domain adaptation, each tumor sample was paired with cell lines within the same cancer type. For example, a patient diagnosed with lung cancer would be paired with lung cancer cell lines. As a result, we created a benchmark dataset to evaluate the performance of the proposed method. The detail of the dataset was listed in Table \ref{tab:dataset}.

\begin{table}[!ht]
    \centering
    \caption{The benchmark dataset for performance evaluation}\label{tab:dataset}
    \begin{threeparttable}
    \begin{tabular}{c|r|r|r|r}
    \hline
        Dataset (domain) & \#CT& \#Sample & \#Drug & \#Pairs  \\ \hline
        GDSC (source) & 23 & 585 & 209 & 123,852  \\ \hline
        PDTC (target) & 1 & 20 & 94 & 1,503  \\ \hline
        TCGA (target) & 23 & 1,249 & 130 & 2,733 \\ \hline
    \end{tabular}
    \begin{tablenotes}
        \footnotesize
        \item[1] \#CT: Number of cancer types.
        \item[2] Sample could be cell lines and PDTXs or patients.
      \end{tablenotes}
   \end{threeparttable}
\end{table}

\subsection{Performance Evaluation on PDTC Drug Response}
The paired dataset of cell lines (source domain) and patient-derived tumor cells (PDTCs) (target domain) was divided into the training set and test set in 8:2 ratio. The performance of the model was evaluated using the Area Under the Receiver Operating Characteristic (AUROC) value. For the cell line drug response, the learned features and predictor yielded an average AUROC value of 0.85 for 94 distinct drugs, indicating that the trained model effectively captured features relevant to drug response. Then we directly aligned the expression profile features of the source domain and the target domain to obtain domain-invariant features, referred as Baseline, which yielded an average AUROC of 0.72 for predicting drug response in the target domain.

Moreover, by integrating the predictions from both the CanCell feature and the TME feature, our Drug2TME model achieved an AUROC value of 0.81, which was significantly better than that achieved by the baseline domain adaptation method. Since Drug2TME is the first method designed to model the effect of the TME feature on drug response, there is no method for straightforward comparison. To benchmark our method, we compared it to CODE-AE \cite{he2022context}, which was the currently state-of-the-art drug response prediction method through cross-domain disentanglement and adaption. CODE-AE built 50 different single drug models and reported AUROC values on PDTC data, from which the mean AUROC 0.7 can be derived. Obviously, our method achieved better performance than CODE-AE on the PDTC data. 

\subsection{Model ablation}
We further performed model ablation experiments to validate the contribution of TME features to precise prediction of drug response. If only the aligned CanCell feature of PDTC samples was taken as input to the cell line predictor for PDTCs drug response prediction, the aligned cancer cell features achieved 0.74 AUROC value. As shown in Table \ref{tab:PDTC}, it was slightly higher than baseline, but significantly lower than our method Drug2TME (CanCell+TME feature). This not only demonstrated the effectiveness of domain-adapted features in predicting drug response in target domain, but also fully confirmed that the decoupled TME features are beneficial for improving the accuracy of drug response prediction for PDTC samples.

\begin{table}[!ht]
    \centering
    \caption{AUROC values achieved by Baseline and Drug2TME on PDTC dataset} \label{tab:PDTC}
    \begin{tabular}{l|c}
    \hline
        Method & PDTCs  \\ \hline
        Baseline & 0.72 \\ \hline
        Drug2TME (CanCell feature) & 0.74  \\ \hline
        Drug2TME (CanCell feature+TME feature)& 0.81  \\ \hline
    \end{tabular}
\end{table}

\subsection{Performance Evaluation on Patient Drug Response}
The primary objective of this study was to validate the capacity of Drug2TME to predict the drug response of tumor patients. Due to the complexity of factors involved in the clinical drug response of tumor patients and the scarcity of clinical data, most existing models considered only a few drugs. Our study was the first to explore the impact of the tumor microenvironment on large-scale drug response. To benchmark the performance of our method, we implemented the Baseline model shown in Figure \ref{fig:baseline}, where the tumor expression profile was aligned to the cell line feature as a whole without feature disentanglement. The domain-adapted feature was then input into the predictor $F_c$ to predict tumor drug response. On the TCGA patient cohort, the Baseline model achieved an AUROC value of 0.58. For comparison, when the first-stage training of Drug2TME was completed, we also tested the decoupled CanCell feature for drug response prediction and obtained only a 0.55 AUROC value. 

\begin{figure}[htbp]
    \begin{minipage}{0.5\textwidth}
        \subfloat[]{\includegraphics[height=4cm]{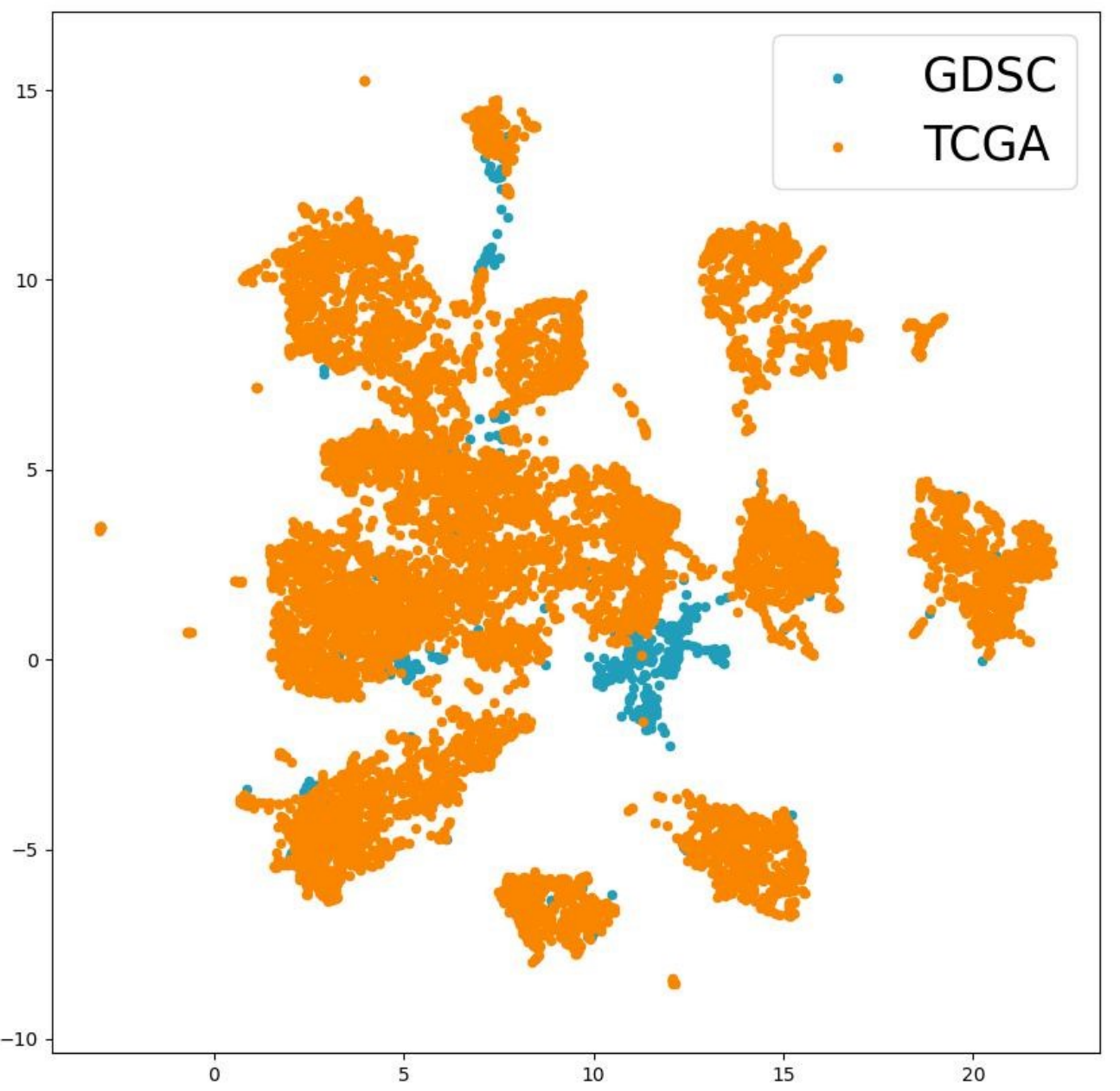}}
        \subfloat[]{\includegraphics[height=4cm]{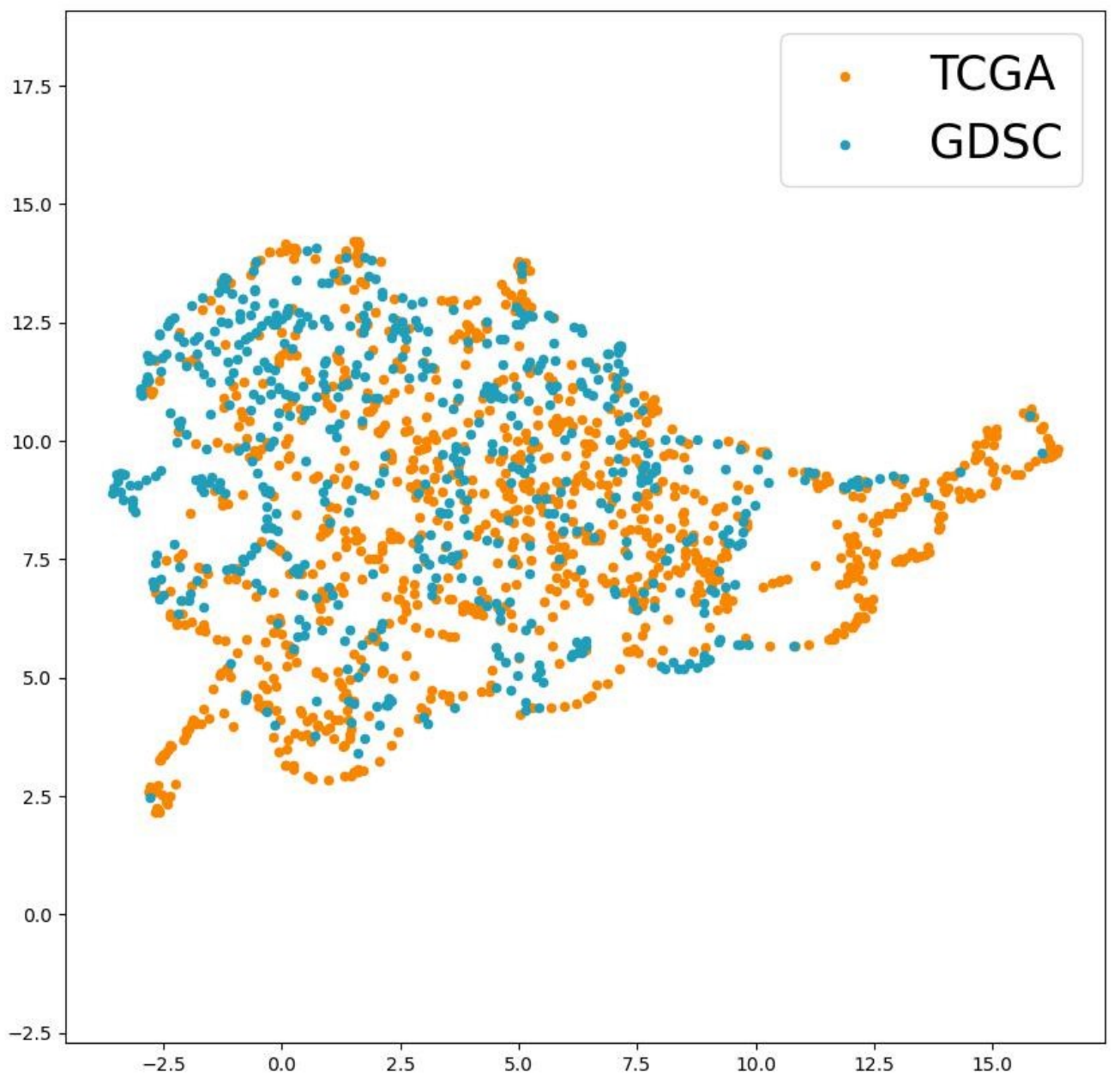}}
    \end{minipage} \\
    \begin{minipage}{0.5\textwidth}
         \subfloat[]{\includegraphics[height=4cm]{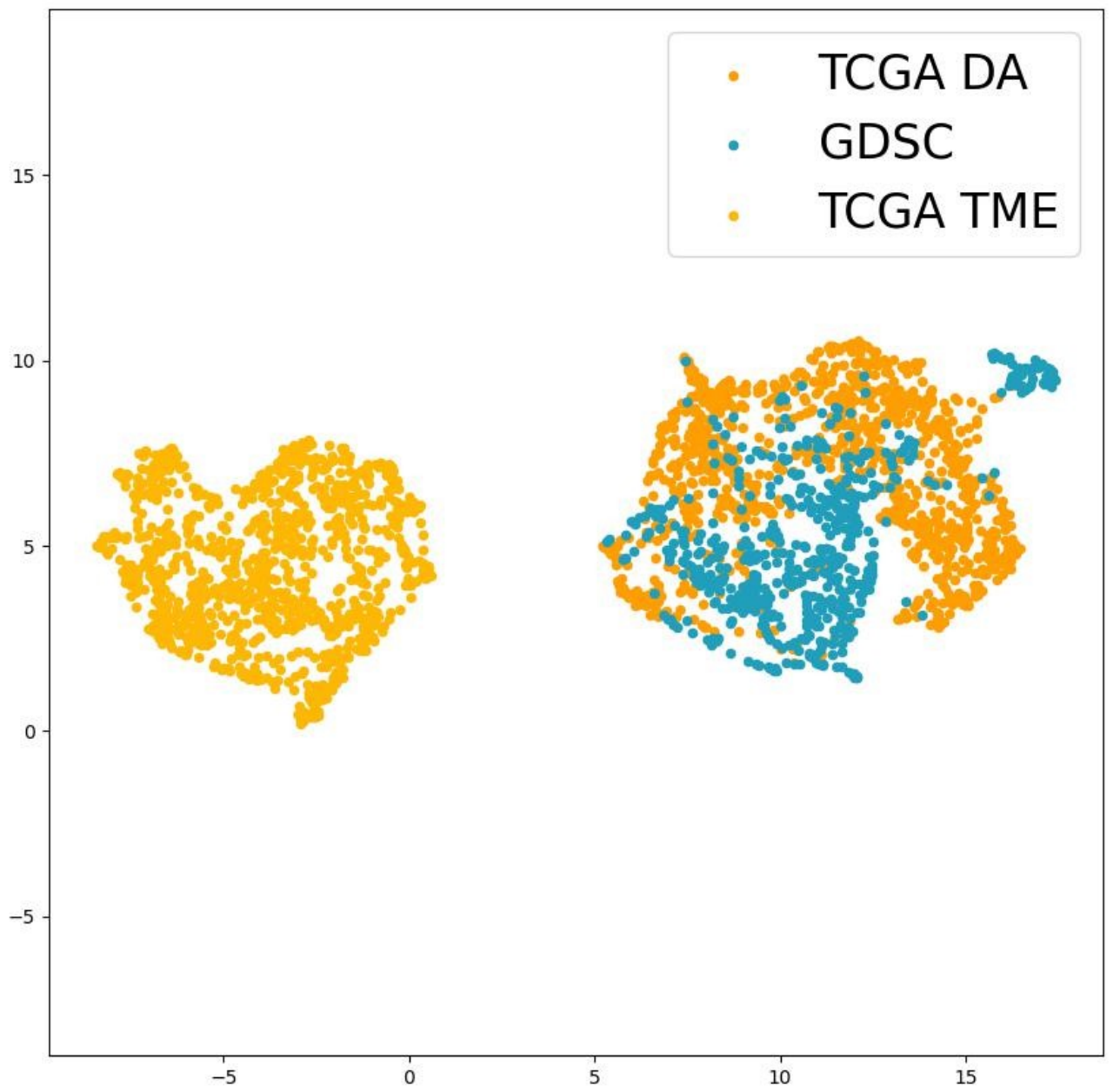}}
        \subfloat[]{\includegraphics[height=4cm]{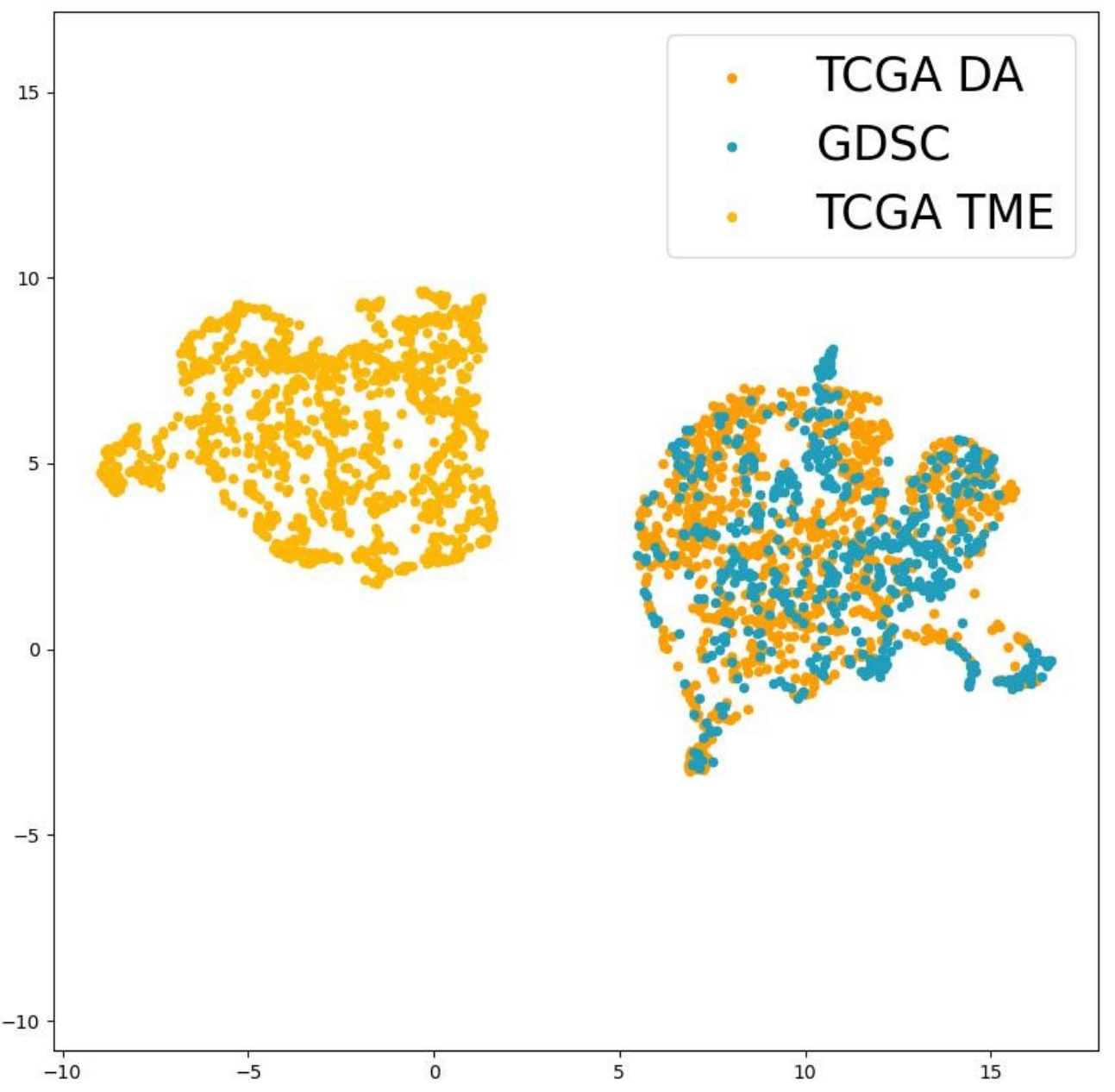}}
    \end{minipage}
    \caption{UMAP visualization of the expression profiles and learned features. (a) The transcriptional profiles of cell lines and tumors. (b) Overall domain-adapted feature between cell lines and tumors. (c) Decoupled tumor features and cell line features. (d) Decoupled tumor features with domain adaptation to cell line features. } \label{fig:umap}
\end{figure}

For intuitive explanation, we used the UMAP tool \cite{UMAP} to perform dimensionality reduction and displayed the data distribution of the transcriptional profiles of cell lines and tumors, domain-adapted features with and without feature disentanglement. As shown in Figure \ref{fig:umap}, the original transcriptional profiles were clustered into a few clusters according to tumor type, with cell lines (source domain) and tumors (target domain) separately distributed. The straightforward domain adaptation between cell lines and tumors led to an aligned but abnormal data distribution (Figure \ref{fig:umap}b). As expected, the decoupled feature of tumors resulted in significant separation between cancer cell and TME features. When tumor feature decoupling and partial domain adaptation were performed simultaneously, the decoupled cancer cell features were well-aligned with cell line features(Figure \ref{fig:umap}c,d).

However, the performance of the baseline model was very low, suggesting that the response of real tumors to drug treatment is influenced not only by cancer cells but also by various factors associated with the tumor microenvironment. Therefore, we restricted the usage of patient labels to test the model performance upon a small set of labeled samples in the target domain. Meanwhile, we paid attention to the balance between the CanCell and TME feature predictions would also be affected by the amount of target-domain labels used for training.Therefore, we incrementally increased the percentage of patient samples used to train the TME predictor from 1\% to 30\% and combined the two predictors to estimate the drug response of the remaining patients. The results, presented in Table \ref{tab:few-shot}, demonstrated that even when only 1\% of patient data is used for training, the model's predictive power improves by 3\%. As the number of patient samples used for training increases, the performance improves significantly. When 30\% of patient samples are used for training, the AUROC value reaches 0.72, representing a substantial improvement over the baseline model. These results indicate that tumor feature decoupling and combination of the impact of both cancer cells and the TME on drug response can achieve better performance.

\begin{table}[!ht]
    \centering
    \caption{Performance increased with percent of TCGA patients used to train TME predictor} \label{tab:few-shot}
    \begin{tabular}{c|c|c}
    \hline
        Percent of training patients & AUROC & AUPRC  \\ \hline
        0\% & 0.55  &  0.62\\ \hline
        1\% & 0.61  &  0.79 \\ \hline
        5\% & 0.66  & 0.78  \\ \hline
        10\% & 0.67 & 0.80  \\ \hline
        20\% & 0.70 & 0.81 \\ \hline
        30\% & 0.72 & 0.80 \\ \hline
    \end{tabular}
\end{table}

\subsection{Quantifying the Impact of TME on Drug Response}
\begin{figure}[htbp]
    \centering
    \centerline{\includegraphics[width=10cm]{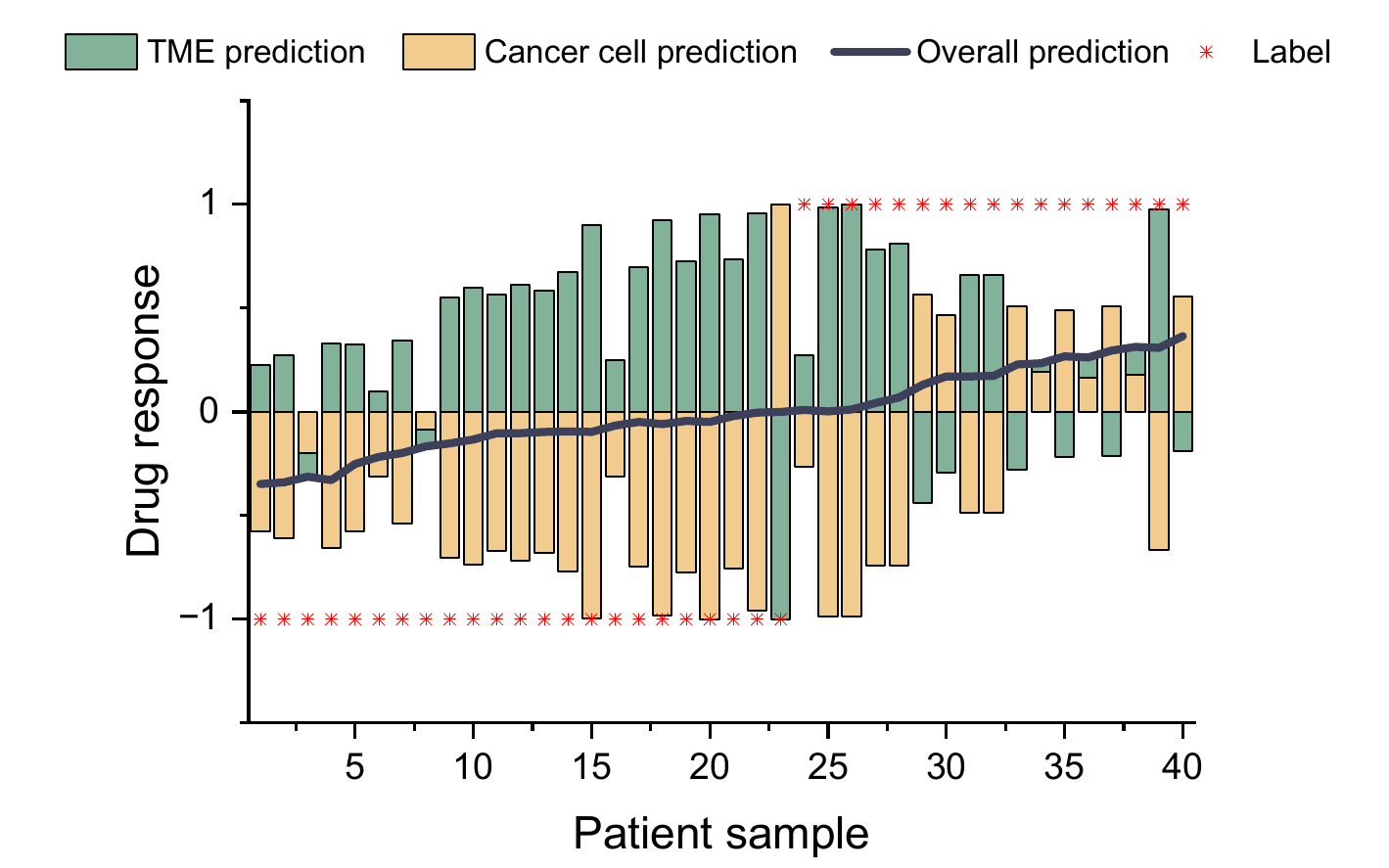}}
    \caption{The predicted scores of 40 TCGA patients derived from cancer cell predictor and TME predictor, respectively.
} \label{fig:bar}
\end{figure}

Upon completion of the second stage of training, the prediction of tumor drug response comprises two components: the cancer cell predictor and the TME predictor. Based on the linear additive assumption, we combined these two components linearly to approximate the actual drug response, enabling us to separate and interpret the impact of the tumor environment on drug response. As illustrated in Figure \ref{fig:bar}, it is evident that for some patients, the predicted score is inaccurate when only cancer cell features are utilized. However, incorporation of the output of the TME predictor brings the final result (blank line) closer to the actual clinical outcome (red asterisk mark).

\section{Conclusion}
In this paper, we propose a domain adaptation network for feature disentanglement to separate representations of cancer cells and the tumor microenvironment of a tumor in patients.  The model is calibrated on a benchmark dataset and demonstrates superior performance in predicting clinical drug response and dissecting the influence of the tumor microenvironment on drug efficacy. To our best knowledge, this is the first study to quantitatively estimate the role of tumor microenvironment in the response to drug treatment.

\bibliographystyle{aaai}
\bibliography{ref}

\end{document}